# IT-map: an Effective Nonlinear Dimensionality Reduction Method for Interactive Clustering

Teng Qiu    ( qiutengcool@163.com)
Yongjie Li*     (liyj@uestc.edu.cn)

Key Laboratory of NeuroInformation, Ministry of Education of China, School of Life Science and Technology, University of Electronic Science and Technology of China, Chengdu, 610054, China
*Corresponding author.

**Abstract**: Scientists in many fields have the common and basic need of dimensionality reduction: visualizing the underlying structure of the massive multivariate data in a low-dimensional space. However, many dimensionality reduction methods confront the so-called "crowding problem" that clusters tend to overlap with each other in the embedding. Previously, researchers expect to avoid that problem and seek to make clusters maximally separated in the embedding. However, the proposed in-tree (IT) based method, called IT-map, allows clusters in the embedding to be locally overlapped, while seeking to make them distinguishable by some small yet key parts. IT-map provides a simple, effective and novel solution to cluster-preserving mapping, which makes it possible to cluster the original data points interactively and thus should be of general meaning in science and engineering.

## 1 Introduction

**Physically inspired IT structure:** in (*1*), we proposed a physically inspired method to organize data points into a sparse yet effective in-tree (IT) structure. In our previous works (*1, 2*), we have shown its potential in cluster analysis. Combinations of the IT structure with the Semi-Supervised learning concept (*3*), Rodriguez and Laio's "Decision Graph" (*4*), and Frey and Dueck's "Affinity Propagation" (AP) (*5*), have resulted in effective cluster analysis methods. For example, based on the IT structure, the application scope of AP was extended from spherical to non-spherical cluster detection (*2*). In this paper, we will show another potential of the IT structure: nonlinear dimensionality reduction, for which an effective combination is made with the "isometric mapping" (Isomap) proposed by Tenenbaum et al (*6*).

**Isomap** is a simple and effective dimensionality reduction method which extends the application scope of multidimensional scaling (MDS) from linear to nonlinear structure. It contains three steps: first construct the K-nearest-neighborhood (KNN) graph, then compute the graph distances (the shortest path distances in the neighborhood graph) and lastly compute the low-dimensional embedding by classical MDS. In effect, the constructed KNN graph for data points is unfolded in the low-dimensional Euclidean space, which is effective especially for preserving in the embedding the topology relationship of data points on manifolds. The crux of the success for Isomap is that it takes as the input for classical MDS the graph distances, instead of the straight-line Euclidian ones, for all pairs of data points.

## 2 Motivation

In short, we would like to replace the KNN graph in Isomap by our physically inspired IT graph, and then, similarly, unfold this IT structure in the low-dimensional space. Consequently, the IT structure constructed directly for the original data points (irrespective its dimensionality and attribute) can always be visualized in the low-dimensional Euclidean space and thus the undesired edges in it can be visualized and removed interactively.

## 3 The proposed method IT-map (3 steps)

**Step 1, construct the IT structure.**

For the original dataset $\chi = \{X_i \mid i = 1, 2, ... , N\}$, we view each point $X_i$ as a node $i$. And the directed edges in IT are defined as follows:

First the potential $P_i$ associated with each point $i$ is computed by

$$P_i = -\sum_{j=1}^{N} e^{-\frac{d_\chi^2(i,j)}{\sigma}} \quad (1)$$

where $\sigma$ is a positive parameter and $d_\chi^2(i,j)$ measures the distance between $X_i$ and $X_j$ by some distance metric (e.g., Euclidean distance). Then, roughly speaking, let any node $i$ "descend" to the nearest node $I_i$, i.e., $I_i$ is roughly[1] defined as

$$I_i = \arg \min_{k \in K_i} d_\chi(i,k), \quad K_i = \{k \mid P_k < P_i\} \quad (2)$$

Here, nodes $i$ and $I_i$ are respectively the start and end node of a directed edge, and "descend" means the potential of the end node should be less than that of the start node. To differentiate with the case after mapping, we denote the IT structure here as $IT_\chi$.

**Step 2, compute the graph distance.**

The graph distance $d_T(i, j)$ of any pair of nodes is defined as the shortest path distance in a "tree" (T) structure, obtained by ignoring[2] the directions of all edges in $IT_\chi$.

We can either use certain algorithm to search the shortest path as what Isomap does, which is, however, time-consuming, or follow a way described in (7) by utilizing the feature of tree structure, that is, the tree structure can be first transformed to an IT structure rooted at node $i$, on which the shortest path between the root node $i$ to any other node $j$ is just the only one path (denoted as $\Gamma_{i,j}$) along the edge directions from node $j$ to $i$, and thus $d_T(i, j)$ is set to the sum of the lengths of the edges on $\Gamma_{i,j}$.

**Step 3, map IT to a low-dimensional Euclidian space R.**

First, compute for each node $i$ its coordinate $Y_i$ in space R using MDS. [3] Based on the input distance matrix $D_T = \{d_T(i,j)\}$, MDS can find a low-dimensional embedding of the original dataset to minimize $\Sigma_{i,j} d_T(i,j) - d_R(i,j)$, where $d_R(i,j) = \|Y_i - Y_j\|$ refers to the Euclidian distance between node $i$ and $j$ in space R. Then, the connection relationships and potential values of the nodes in $IT_\chi$ are all inherited to the corresponding nodes embedded in space R.

---

[1] See a more elaborate definition in ref. 1.
[2] In practice, this means that the end node of any directed edge is connected to its start node, equivalent to transforming all directed edges into undirected ones
[3] We used non-classic MDS Matlab code as in Fig2. A and C, the results of which can be slightly better than MDS. However, non-classic MDS code sometimes has the risk of being hard to converge sometimes. So we used the classic MDS for most datasets.

Consequently, $IT_\chi$ (Fig. 1B) is mapped into space R, the result denoted as $IT_R$ (Fig. 1C). However, we prefer to visualize $IT_R$ in the RP space (the result denoted as $IT_{RP}$), where potentials on nodes are shown by an additional dimensionality (potential-axis) as in Fig. 1D. This helps users better visualize the clusters, especially for the undesired edges.

## 4 Experiments

We tested the power of IT-map on several 2-dimensional datasets from (4, 8, 9). Figure 2 shows the $IT_{RP}$ structures (the dimensionality of R is 1) after mapping, where clusters, together with the undesired edges across them, are distinguishable, proving that IT-map succeeds in these mapping tasks. These friendly and beautiful visualized results (*"a group of penguins spitting water to each other"*) are so useful that it is possible for users to make reliable cluster analysis just by simple interactive operations (recorded by the red crosses in Fig. 2). The edges closest[4] to the red crosses will be determined as the undesired ones and removed, and consequently, the clustering assignments for the original data points are shown in the upper-right corners, which is quite consistent with visual perception.

In fig. 3, we tested a high-dimensional synthetic dataset (N=1024 vectors, d = 32 dimensionalities, M=16 Gaussian clusters, numerical attribute) (10). The clustering assignments fully match with the ground truth, i.e., 1024 vectors in the original dataset are assigned into 16 clusters without any error. In fig. 4, we tested the high-dimensional real-world "mushroom dataset[5]" (N = 8124, d=22, M = 2 classes[6], character attribute). 8124 mushrooms are assigned into 24 clusters with an error rate of 0.0039, which is much better than the interactive clustering based on ISOMAP mapping result reported in our first paper (*1*).

## 5 Discussion and conclusion

**Comparing with other visualization methods:** visualization is required in diverse domains and many methods as reviewed in (11) have been proposed. Most of these methods as Chernoff faces (12) and pixel-based techniques (13), require the interpretation of user to the symbols in the embedding and thus have limitation when dealing with large volume of dataset (14). In contrast, our method not only represents high-dimensional dataset simply as points in the low-dimensional space, but also organizes those points efficiently in an IT structure (a special directed tree). ***These***

---

[4] See details for the method of how the undesired edges are identified by these red crosses in ref. 1

[5] From http://archive.ics.uci.edu/ml/

[6] Note that these mushrooms are classified by people into 2 classes as poisonous or edible, whereas the underlying number of clusters should be much more than that. The true number of clusters is more likely to be 23 as revealed in our another algorithm(G-AP, ref. 2).

*help both visualization and further interactive operations as in this paper.*

**Comparing with other dimensionality reduction methods:** it is hard (15, 16) for most dimensionality reduction methods particularly as principal components analysis (PCA) (17), classical multidimensional scaling (MDS) (18), Sammon mapping (19), locally linear embedding (LLE) (20), stochastic neighbor embedding (SNE) (21), Maximum Variance Unfolding (MVU) (22) and ISOMAP, to map data points into a low-dimensional space while preserving the underlying cluster structure due to the so-called "crowding problem"[7] (14). Previous methods as tree preserving embedding (TPE) (16) and the variants (14, 15, 23, 24) of SNE try to maximally alleviate the crowding or overlapping phenomenon of clusters. In contrast, IT-map allows the local overlapping phenomenon to happen[8] as in Fig. 2 and seeks to make clusters distinguished by some salient local features instead of a fully separated image in the low-dimensional space. *This new and reasonable strategy, quite similar[9] to that utilized by card players, provides a simple yet effective solution and should be of general meaning.*

**Comparing with KNN- and MST-based mapping:** for one thing, IT-map is an effective application of the physically inspired IT structure in dimensionality reduction. For another, like the minimal spanning tree (MST) (8) based mapping (25-27), IT-map (IT-based mapping) is also a variant of Isomap (KNN-based mapping) and inherits the virtue of Isomap in preserving the graph distance (step 2~3) while "unfolding" the graph in the low-dimensional Euclidean space. However, *IT-map is a more effective nonlinear dimensionality reduction method for interactive clustering*, since the physically inspired IT structure has some salient advantages over traditional graph structures as KNN and MST:

(**i**) Compared with KNN and MST, *the main difference or advantage for the IT structure* is that the neighbor of each point is not constrained in a local area[10], especially for the local extreme points in terms of potential variable. Consequently, the connections between clusters are generally much longer (see a comparison of MST and IT in Fig. S11 and Fig. 2A in ref. 1). Therefore, according to step 2, the paths $\Gamma_{i,j}$ connecting the nodes between clusters are generally longer than that within clusters, and according to step 3, more compact (small distance within clusters) and distinguishable (large distance between clusters) clusters generally present in the embedding, which can facilitate the cluster analysis (Fig. 2). (**ii**) Compared with KNN and MST, the nodes of the IT structure are also weighted with potential values, which provides an additional yet useful reference or dimensionality to make clusters separable (Fig. 1D and Fig. 2). (**iii**) Unlike KNN and MST, the directed characteristic of the IT structure makes it sparser and also brings convenience in identifying the undesired edges and searching the root nodes in interactive clustering. Compared with KNN, the tree-shaped feature for IT (also for MST) makes each removing of one edge surely divide the graph into two separate parts, which makes it convenient for the "divide and conquer" strategy introduced next.

---

[7] Clusters are overlapped mutually

[8] In fact, we can interactively drag the overlapped clusters apart in the low-dimensional space as shown in Fig. 3.

[9] It is easy for us to place several cards fully separated on the table, yet not easy if one intends to complete the same task in one hand. However, it is still not too hard for those card players, since they choose a strategy similar as ours by distinguishing those cards just by the characters or numbers in the upper-left corners of the cards and allowing the rest parts overlapped with each other.

[10] This difference is due to the potential variable introduced in Eq. 2. Therefore, strictly speaking, this IT structure is not a neighbor graph like KNN and MST.

**Problems and solutions:** although we don't seek to avoid the "crowding problem", sometimes it is still expected to slightly improve it by some post-processing due to two main problems that may be encountered in practice: (**i**) sometimes the embedded structure is too crowded for user to easily spot the undesired edges; (**ii**) the current parameter σ is not the one best fitted for the test dataset, or for some dataset, there is not at all any single value well enough to lead to an optimal result.

One can utilize the "***divide and conquer***" strategy, which can be further decomposed into two components: the "divide" and "conquer" strategies. (**a**) The 1st problem can be solved by the "divide" strategy by rerunning step 3 to a subset of the distance matrix (corresponding to a sub-dataset denoted as χ'), as illustrated in Fig. S1. Namely, instead of expecting the dimensionality of space, we can downsize the number of clusters in the original IT structure, since the burden of step 3 for IT-map is mainly derived from the number of clusters. (**b**) If the "divide" strategy doesn't work such as when encountering the 2nd problem, one can further use the "conquer" strategy by adjusting the value of the parameter to χ' and rerunning steps 1~3, until a salient result presents. In order to be more efficient and reliable, some information as some constraints (e.g., the "must-link" or "cannot-link" constraint[11]) or several labels in semi-supervised learning (3) are always welcome to provide some references. Figure. S2 illustrates with one simple example the role of several labels when "conquer" strategy is used. [12]

**The Meaning of IT-map**: we believe IT-map can boost the interactive clustering method proposed in (*1*) to have a broad meaning. As stated by Shneiderman (28): *"A more effective approach will be to put human users in control, since they can often identify patterns that machines cannot...automated analyses can work for well-understood data, but visualizations increase the efficacy of experts in frontier topics, where big breakthroughs happen."*

**References**
1. Qiu T, Yang K, Li C, & Li Y (2014) A Physically Inspired Clustering Algorithm: to Evolve Like Particles. *arXiv preprint arXiv:1412.5902*.

---

[11] The "must-link" denotes that some nodes should exist in one cluster, opposite to the case for the "cannot-link".
[12] One may think if one label is given in each cluster, even Kmeans can work well. However, (i) this illustration can be generalized to "must-link" or "cannot-link" information, which is impractical for Kmeans; (ii) the test data can also be non-spherical or character-attributed, also impractical for Kmeans; (iii) the label in each cluster can be generally any member, which is also impractical for Kmeans due to its initialization sensitivity; (iv) more constraint means more reliable cutting behavior, which is not necessarily the case for Kmeans.

2. Qiu T & Li Y (2015) A Generalized Affinity Propagation Clustering Algorithm for Nonspherical Cluster Discovery. *arXiv preprint arXiv:1501.04318*.
3. Chapelle O, Schölkopf B, & Zien A (*Semi-supervised learning (MIT press, Cambridge, MA, 2006)*.
4. Rodriguez A & Laio A (2014) Clustering by fast search and find of density peaks. *Science* 344(6191):1492-1496.
5. Frey BJ & Dueck D (2007) Clustering by passing messages between data points. *Science* 315(5814):972-976.
6. Tenenbaum JB, De Silva V, & Langford JC (2000) A global geometric framework for nonlinear dimensionality reduction. *Science* 290(5500):2319-2323.
7. Qiu T & Li Y (2014) An Effective Semi-supervised Divisive Clustering Algorithm. *arXiv preprint arXiv:1412.7625*.
8. Zahn CT (1971) Graph-theoretical methods for detecting and describing gestalt clusters. *IEEE Trans. Comput.* 100(1):68-86.
9. Fränti P & Virmajoki O (2006) Iterative shrinking method for clustering problems. *Pattern Recognit.* 39(5):761-775.
10. Franti P, Virmajoki O, & Hautamaki V (2006) Fast agglomerative clustering using a k-nearest neighbor graph. *Pattern Analysis and Machine Intelligence, IEEE Transactions on* 28(11):1875-1881.
11. De Oliveira MCF & Levkowitz H (2003) From visual data exploration to visual data mining: a survey. *Visualization and Computer Graphics, IEEE Transactions on* 9(3):378-394.
12. Chernoff H (1973) The use of faces to represent points in k-dimensional space graphically. *Journal of the American statistical association* 68(342):361-368.
13. Keim DA (2000) Designing pixel-oriented visualization techniques: Theory and applications. *Visualization and Computer Graphics, IEEE Transactions on* 6(1):59-78.
14. Van der Maaten L & Hinton G (2008) Visualizing data using t-SNE. *Journal of Machine Learning Research* 9(2579-2605):85.
15. Venna J, Peltonen J, Nybo K, Aidos H, & Kaski S (2010) Information retrieval perspective to nonlinear dimensionality reduction for data visualization. *The Journal of Machine Learning Research* 11:451-490.
16. Shieh AD, Hashimoto TB, & Airoldi EM (2011) Tree preserving embedding. *Proc. Natl. Acad. Sci. U.S.A.*
17. Hotelling H (1933) Analysis of a complex of statistical variables into principal components. *Journal of educational psychology* 24(6):417.
18. Torgerson WS (1952) Multidimensional scaling: I. Theory and method. *Psychometrika* 17(4):401-419.
19. Sammon JW (1969) A nonlinear mapping for data structure analysis. *IEEE Trans. Comput.* 18(5):401-409.
20. Roweis ST & Saul LK (2000) Nonlinear dimensionality reduction by locally linear embedding. *Science* 290(5500):2323-2326.
21. Hinton GE & Roweis ST (2002) Stochastic neighbor embedding. *Advances in neural information processing systems*, pp 833-840.
22. Weinberger KQ, Sha F, & Saul LK (2004) Learning a kernel matrix for nonlinear dimensionality reduction. *Proceedings of the twenty-first international conference on Machine learning*, (ACM), p 106.

23. Cook J, Sutskever I, Mnih A, & Hinton GE (2007) Visualizing similarity data with a mixture of maps. *International Conference on Artificial Intelligence and Statistics*, pp 67-74.
24. Carreira-Perpinán MA (2010) The Elastic Embedding Algorithm for Dimensionality Reduction. *ICML*, pp 167-174.
25. Cannistraci CV, Ravasi T, Montevecchi FM, Ideker T, & Alessio M (2010) Nonlinear dimension reduction and clustering by Minimum Curvilinearity unfold neuropathic pain and tissue embryological classes. *Bioinformatics* 26(18):i531-i539.
26. Cannistraci CV, Alanis-Lobato G, & Ravasi T (2013) Minimum curvilinearity to enhance topological prediction of protein interactions by network embedding. *Bioinformatics* 29(13):i199-i209.
27. Alanis-Lobato G, Cannistraci CV, Eriksson A, Manica A, & Ravasi T (2015) Highlighting nonlinear patterns in population genetics datasets. *Scientific reports* 5.
28. Shneiderman B (2014) The big picture for big data: visualization. *Science* 343(6172):730.

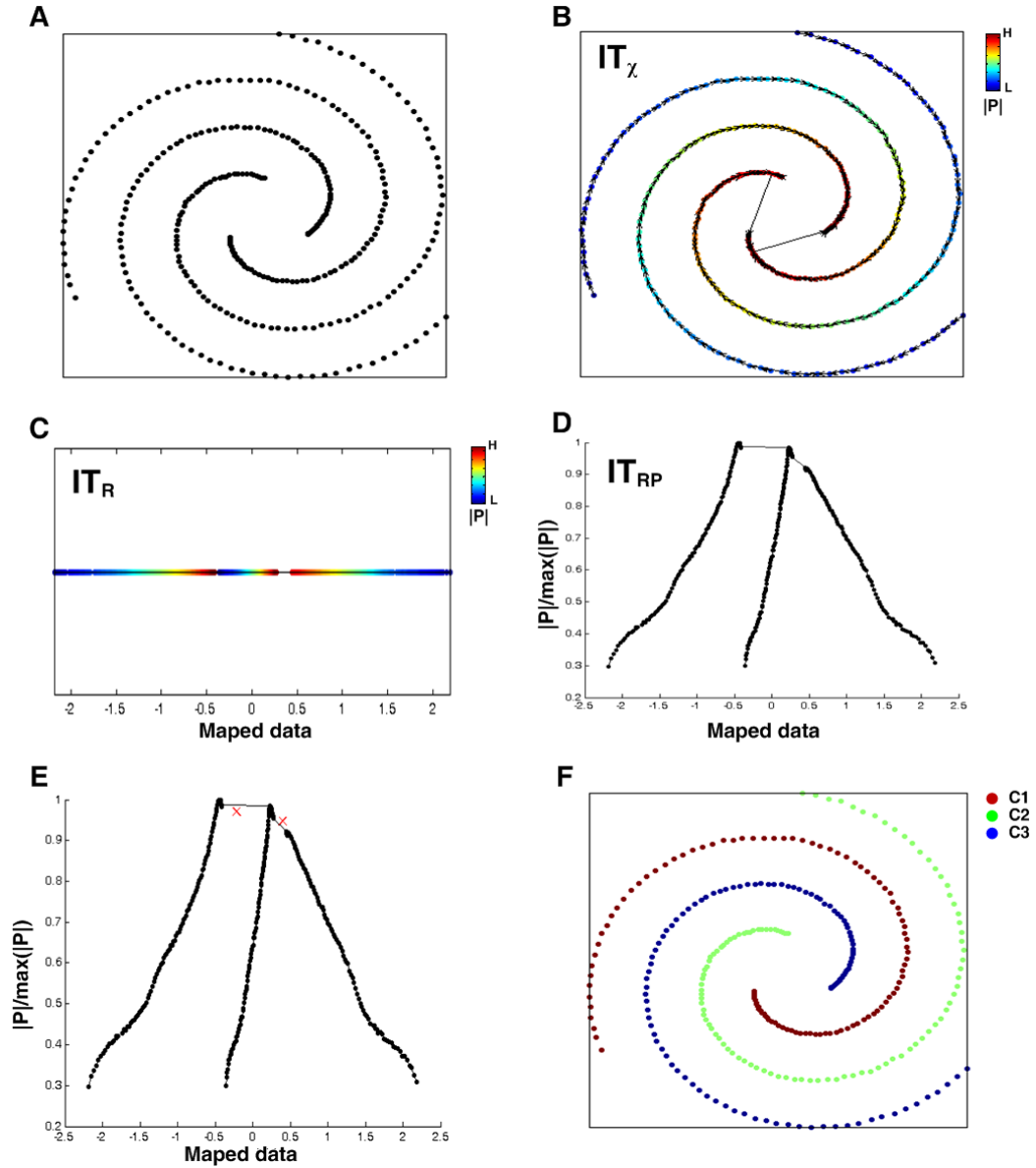

**Fig. 1. An illustration for the proposed nonlinear dimensionality reduction (A~D) and its application in interactive clustering (E~F)**. (**A**) A 2-dimensional (2D) spiral dataset. (**B**) IT$_\chi$, the IT ($\sigma$ = 4) structure for the original dataset. Different colors on points represent different potentials. The redder of a node, the higher the magnitude (|P|) of the potential is. (**C**) IT$_R$, the result after mapping IT$_\chi$ into the 1D Euclidian space R. Horizontal axis shows the coordinates of all points in space R.. (**D**) IT$_{RP}$, another representation of the IT structure in (C). Here the magnitudes (normalized) of potentials are shown in an additional dimension (vertical axis). (**E**) Interactive clustering. Red crosses record user's operations. The edges closest to them will be determined as the undesired edges and the corresponding edges in (B) will be removed. (**F**) Clustering result. This result is quite consistent with visual perception in (A), which also indirectly indicates that the undesired edges determined in (E) just correspond to those undesired ones in (B).

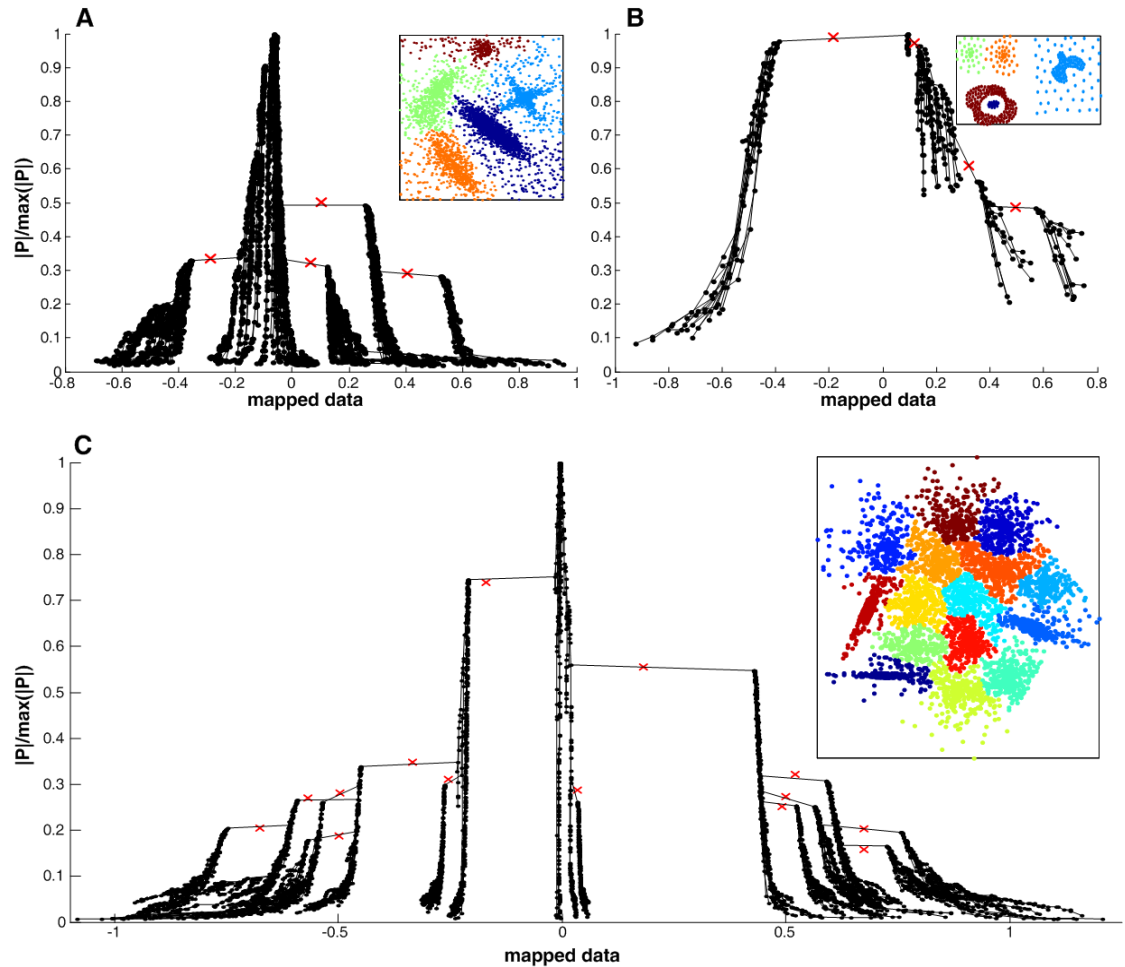

**Fig. 2. The embedded IT structures and the interactive clustering results of three datasets.** From A to C, σ = 0.01, 2, 10000, respectively. Red crosses record users' interactive operations. Each embedded IT image can be viewed as an image of "a group of penguins spitting water to each other". Each penguin represents a cluster. For each cluster, the spiky side represents its head. Water (denoting the undesired edges) is spitted from its mouth. The other side represents its tail. Usually, the tails of different clusters (or penguins) are heavily overlapped. However, since the heads of the penguins (or clusters) are visually distinguishable, the edges started from the head are also distinguishable. Therefore, for cluster analysis, we only need to select as the undesired edges the ones that are started from the mouth-like parts.

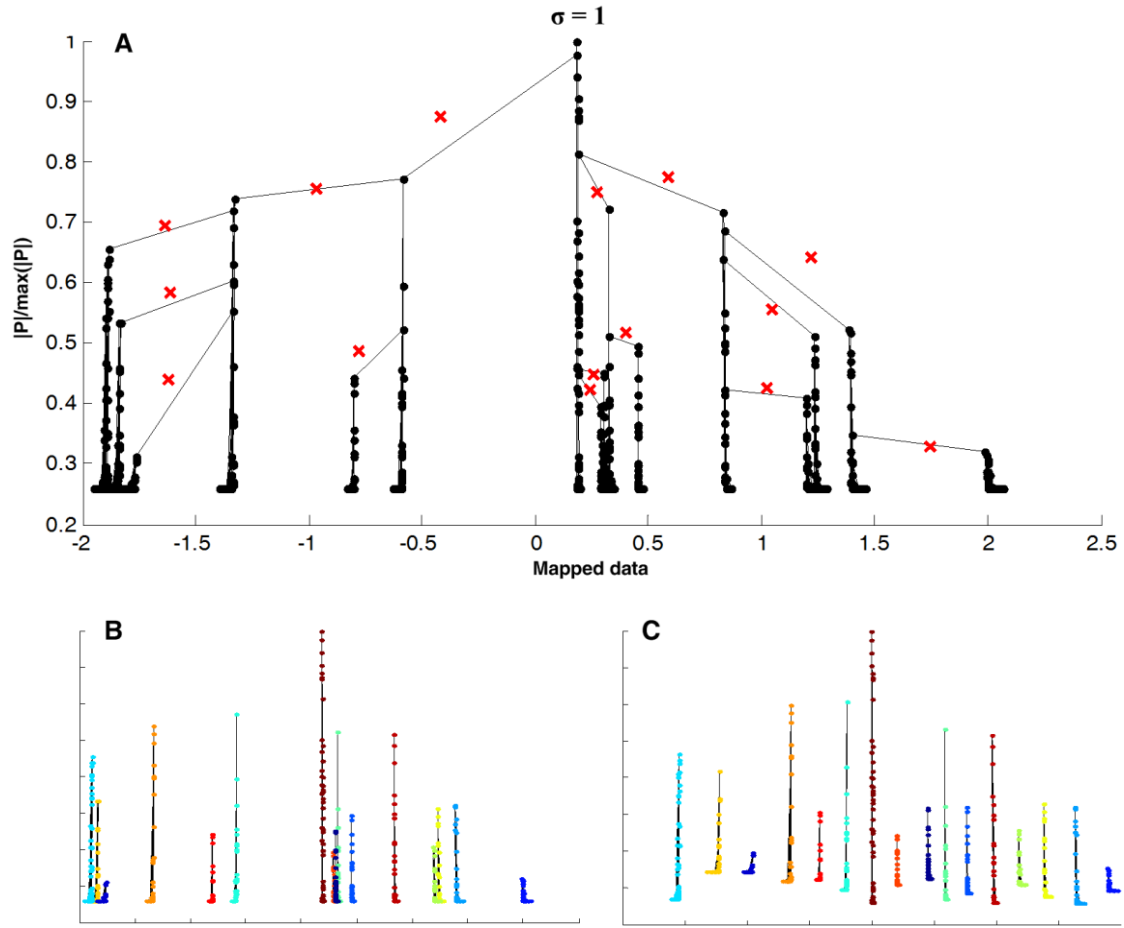

**Fig. 3. Embedded result of high-dimensional dataset (N=1024, d=32, M=16 classes, numerical attributes)**. Clustering assignments to original data: 16 clustering, error rate: 0. (**A**) the IT$_{RP}$ representation of embedding. (**B**) The result after removing the undesired edges identified by the red crosses in (A). Several independent sub-graphs are obtained. Colors on points denote the Ground Truth (true classes) of the corresponding points in original dataset. *Since the sub-graphs are independent, we can drag apart the overlapped sub-graphs to view the result more clearly.* The result is shown in (C), where it shows that points in each sub-IT structure are of the same color, representing that the clustering assignments to the corresponding data points in the original dataset have no error clustering assignment.

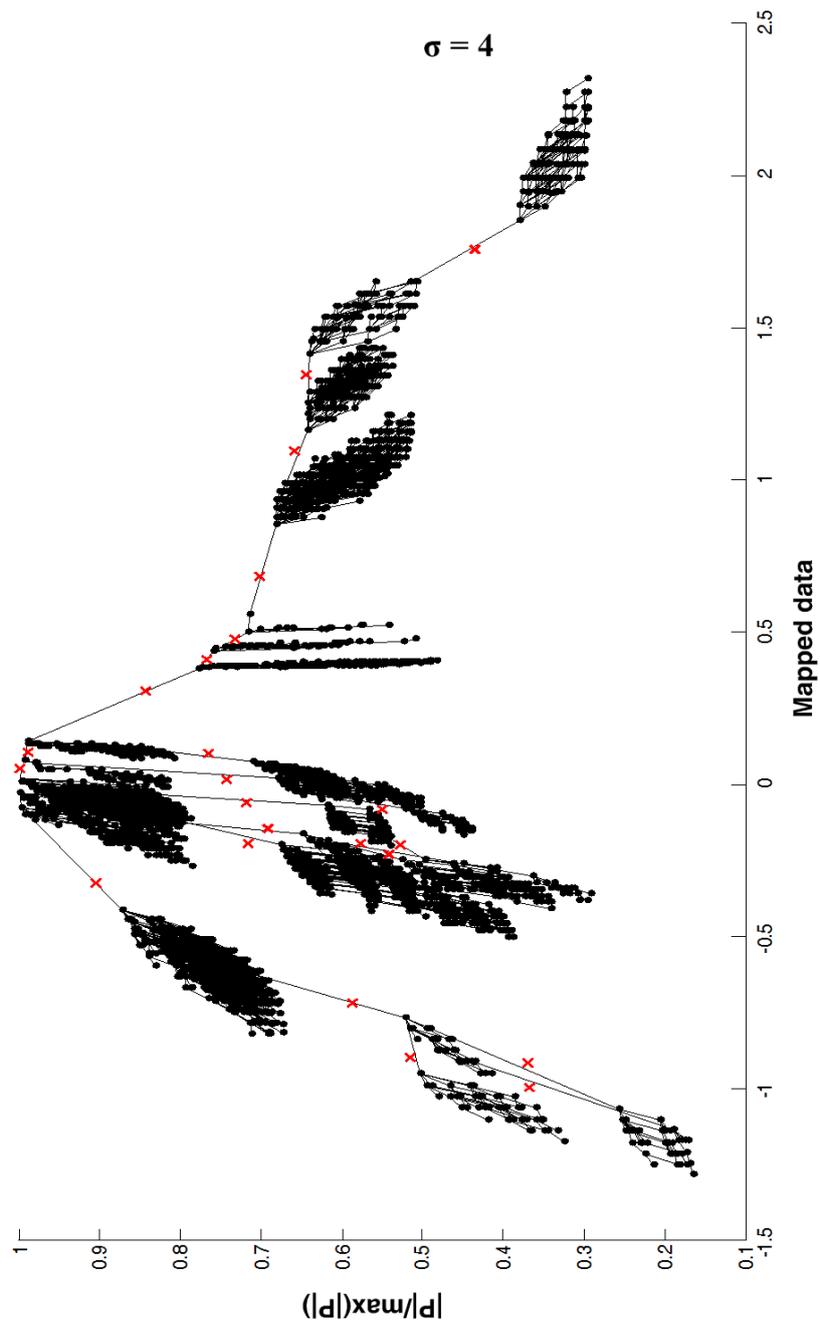

**Fig. 4. Embedded result of the high-dimensional mushroom dataset (N = 8124, d = 22, M = 2 classes, character attributes).** Clustering assignments to original data: 24 clustering, error rate: 0.0039.

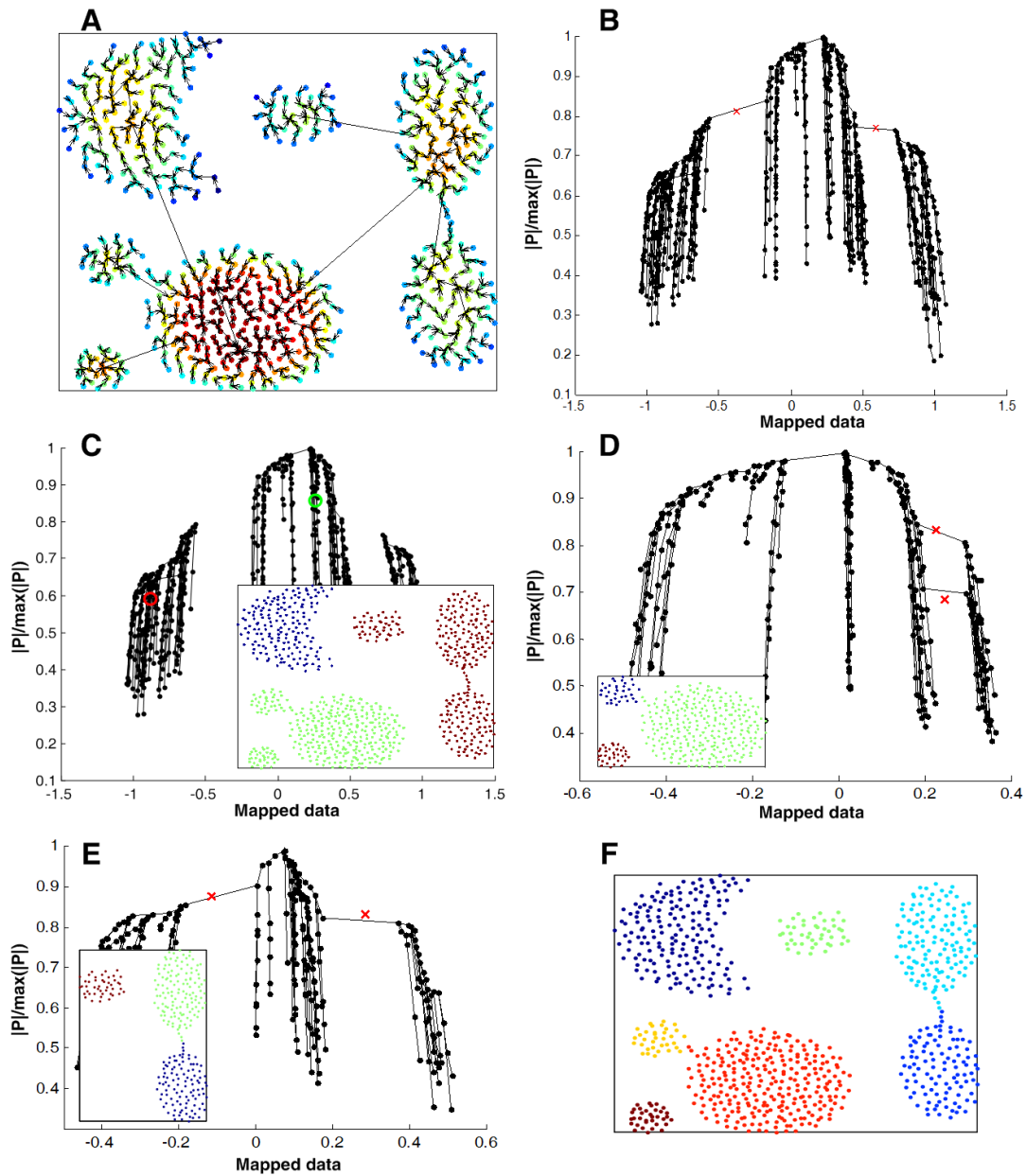

**Fig. S1. An illustration for the "divide" strategy.** (**A**) The original 2D dataset and the its IT structure (σ = 1). Colors on points denotes different potential values. (**B**) $IT_{RP}$. (**C**) The result after removing the undesired edges identified by red crosses in (B) and the corresponding clustering results (bottom-left) of the original dataset. Three clusters are obtained, since only three independent sub-ITs are obtained. Two of clusters have error assignments. In order to make a further partitioning, two of these sub-structures, identified by the red and green circles respectively, are further observed in (**D**) and (**E**) with the clustering results corresponding the subset of the original dataset shown in the bottom-left. The error clusters are further divided in (D) and (E). (**F**) The ultimate result after integrating the results in (C~E), where the clustering assignment is in line with visual perception.

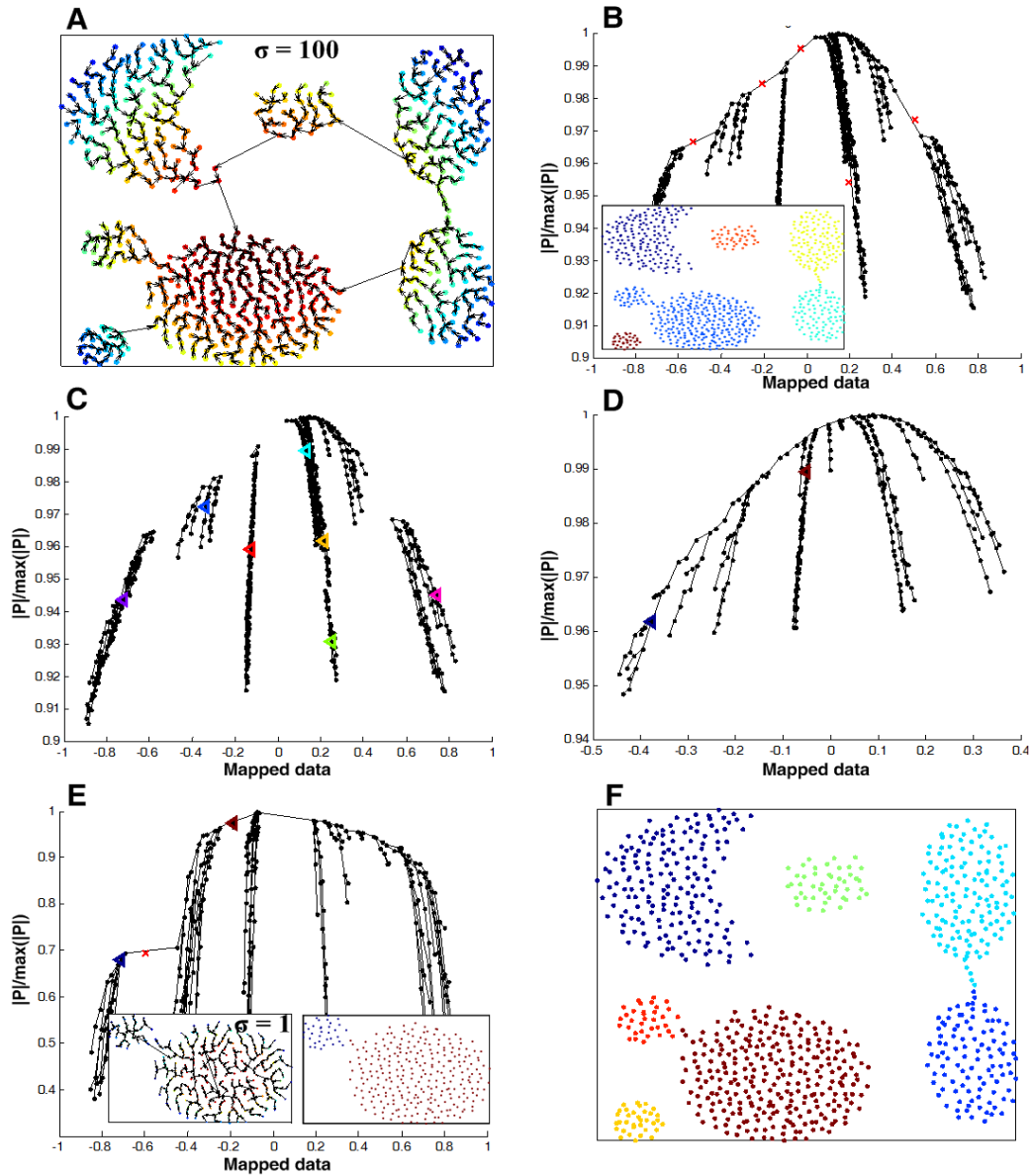

**Fig. S2. An illustration for the "conquer" strategy by adjusting the parameter σ for the sub-dataset.** (**A**) The IT structure (σ = 100). Different colors on points denote different potential values. The redder, the lower of the potential. (**B**) $IT_{RP}$. Bottom-left: clustering results of the original dataset after removing the undesired edges identified by the red crosses. (**C**) The result after removing the undesired edges in (B). Seven colored (denoting different categories) triangulars correspond to the labeled data in the original dataset. One out of six independent sub-ITs falsely contains two different labeled points (or triangulars), corresponding to the cluster with false clustering assignments in (B). We denote the problematic sub-IT as IT' and the corresponding sub-dataset as χ'. (**D**) The result after independently expanding IT'. No explicit clue is shown, either. (**E**) The $IT_{RP}$ of sub-dataset χ' after adjusting the parameter (σ = 1). Salient undesired edge is shown and the labeled points (triangulars) are rightly distributed on the two sides of the undesired edges. Bottom-left and Bottom-right: the IT structure and clustering result of χ'. (**F**) The ultimate result of the original dataset after integrating the results in (B) and (E).